\documentclass[a4paper,11pt,
abstract=on
]{scrartcl}
\usepackage{amsmath,amsfonts}
\usepackage{algorithm}
\usepackage{array}
\usepackage{textcomp}
\usepackage{stfloats}
\usepackage{url}
\usepackage{verbatim}
\usepackage{graphicx}
\usepackage{cite}
\usepackage{amssymb,siunitx,commath}
\usepackage{amsthm}
\usepackage{mathtools}
\mathtoolsset{showonlyrefs}
\usepackage{subcaption}
\usepackage{algpseudocode}
\usepackage{multirow}
\usepackage{xcolor}
\usepackage{hyperref}

\DeclareMathOperator*{\argmax}{arg\,max}
\DeclareMathOperator*{\argmin}{arg\,min}
\DeclareMathOperator*{\blur}{blur}

\newcommand{\R}{\mathbb{R}}
\newcommand{\C}{\mathbb{C}}

\newcommand{\tT}{\mathrm{T}}

\theoremstyle{plain}
\newtheorem{lemma}{Lemma}
\newtheorem{theorem}[lemma]{Theorem}

\newtheorem{remark}[lemma]{Remark}
\theoremstyle{definition}

\begin{document}
\title{Wasserstein Patch Prior for Image Superresolution}
\author{Johannes Hertrich\thanks{
TU Berlin,
Stra{\ss}e des 17. Juni 136, 
D-10623 Berlin, Germany,
j.hertrich@math.tu-berlin.de.}\and
Antoine Houdard\thanks{
Ubisoft La Forge,
Bordeaux, France, antoine.houdard@ubisoft.com}\and 
 Claudia Redenbach\thanks{Technische Universit{\"a}t Kaiserslautern, Mathematics Department, 67663 Kaiserslautern, Germany, redenbach@mathematik.uni-kl.de}
}

\maketitle

\begin{abstract}
In this paper, we introduce a Wasserstein patch prior for superresolution of two- and three-dimensional images.
In addition to the low-resolution observation, we assume that we are given a reference image which has a similar patch distribution as the ground truth of the reconstruction.
This assumption can e.g.\ be fulfilled when working with texture images or homogeneous material data.
Then, the proposed regularizer penalizes the $W_2$-distance of the patch distribution of the reconstruction to the patch distribution of the reference image at different scales.
We demonstrate the performance of the proposed method by applying it to two- and three-dimensional images of materials' microstructures.
\end{abstract}

\section{Introduction}
Superresolution is the task to reconstruct a high-resolution image based on a low-resolution observation. Many superresolution approaches found in the literature focus on natural 2D images. However, problems due to low resolution may also arise in processing and analysis of (3D) images of materials' microstructures. For instance, selecting the voxel size in micro computed tomography requires a trade-off between representativity of the imaged volume and ability to reconstruct fine structure details such as thin fibres or cracks \cite{Schladitz}. In serial sectional imaging techniques such as scanning electron microscopy coupled with focused ion beam milling (FIB-SEM), limiting the processing time may require to choose the voxel size larger than desired, in particular in the slicing direction. The resulting anisotropic voxels induce an artificial anisotropy in the data which influences the results of a quantitative image analysis \cite{ROLDAN2021113291}. In these cases, application of a superresolution approach may help to overcome limitations due to the imaging setup. However, the reconstruction quality must be validated critically to ensure that geometric or material characteristics obtained from the reconstructed image are reliable.

From a mathematical point of view, image superresolution belongs to the class of inverse problems. 
More precisely, we assume that the high-resolution image $x\in\R^{n}$ and the low-resolution observation $y\in\R^d$ are
related by
$$
y=f(x)+\eta,
$$
where $f\colon\R^{n}\to\R^d$ is a (not necessarily linear) operator and $\eta$ is some noise.
For inverse problems (including superresolution), the operator $f$ is usually differentiable but ill-posed or not injective.
Thus, reconstructing the true image $x$ from a given observation $y$ is difficult and we have to use some prior information.
This is classically done by minimizing a functional
$$
\mathcal J (x)=d(x,y)+\lambda R(x),
$$
where $d$ is a data fidelity term, which ensures that the distance of $f(x)$ and $y$ is small and $R$ is a regularizer, which 
incorporates the prior information. The hyper parameter $\lambda$ serves as a weighting between both terms.
If $\eta$ is Gaussian noise, then the data fidelity term is usually set to $d(x,y)=\tfrac12\|f(x)-y\|^2$.
During the last decades a wide range of regularizers was proposed for different inverse problems.

Simple regularizers, as the total variation (TV) \cite{ROF1992}, can be used very generally for a large number of problems, 
but have usually a weaker performance than reconstruction methods which are adapted to specific problems and
the special structure of the images under consideration.
Therefore, many reconstruction methods for inverse problems make use of the similarities of small ${p\times p}$ 
patches within natural images or images with special structure to derive powerful methods for denoising \cite{BCM2005,HBD2018,LBM2013}, noise level estimation \cite{HHLS2021,SDA2015} or 
superresolution \cite{HNABBSS2020,SJ2016}.

Here, we particularly focus on so-called patch-priors \cite{PDDN2019,ZW2011}. 
Patch priors are regularizers which depend on the patch distribution of $x$, i.e.\ they are of the form
$$
R(x)=\tilde R((P_i x)_{i=1}^N),
$$
where the values $P_i x$ are $p\times p$ patches from $x$ and the operator $P_i$ extracts the $i$-th patch of $x$.

Using the idea that a texture or texture-like image can be represented by its patch distribution, the authors of \cite{EL1999, GRGH2017, HLPR2021, LB2019} proposed to synthesize textures by minimizing the Wasserstein distance of the patch-distribution of the synthesized image to the patch-distribution of some reference image.

Inspired by \cite{HLPR2021}, we propose in this paper to use the Wasserstein-$2$ distance from the patch distribution of the reconstruction to the patch distribution of our reference image as a patch prior for superresolution.
We describe, how to minimize the arising objective functional and provide numerical examples on two- and three-dimensional material data.

We are aware of the fact that many state-of-the-art superresolution methods are nowadays based on machine learning with deep neural networks, see e.g. \cite{DLHT2015, RIM2017,SHCSFN2021, WCH2020,ZTKZF2018}. 
However, most of these methods are trained on registered pairs of high- and low-resolution images.
Unfortunately, such data is often not available in practice. For instance, FIB-SEM imaging is a destructive process which makes it impossible to image the same sample at various voxel sizes.
Instead of employing a large data base of high- and low-resolution image pairs, 
the proposed method needs only a single high-resolution reference image, which is not required to have a low-resolution counterpart.
For example, the high-resolution reference image could show a different region of the same texture or material than the low-resolution observation.
There are only very few neural network based approaches with similar assumptions. Examples are the deep image prior \cite{UVL2018}
and Plug-and-Play (PnP) methods \cite{SVWB2016,VBW2013}. We compare the proposed method with both in Section~\ref{sec_numerics}.

Note that even though we focus on superresolution, the proposed method works theoretically for 
general inverse problems on images.\\

The paper is structured as follows: 
In Section~\ref{sec_prelim} we briefly recall the definition and notations of optimal transport and Wasserstein distances. 
Afterwards, in Section~\ref{sec_wasserstein_patch_prior}, we introduce the proposed regularizer and describe how the functional consisting of data fidelity term and regularizer can be minimized.
Finally, we provide numerical examples in Section~\ref{sec_numerics}.
Conclusions are drawn in Section~\ref{sec_conclustions}.

\section{Preliminaries: Optimal Transport and Wasserstein Distances}\label{sec_prelim}

\noindent
In this section, we briefly recall the definitions and theorems about optimal transport cost and Wasserstein distances that we will use in this paper. For the interested reader, we refer to the existing literature on the subject~\cite{AG2013,PC2019,V2009,santambrogio2015optimal}.

Let $\mu$ and $\nu$ be two measures on $\R^d$. We denote by $\Pi(\mu,\nu)$ the set of all measures $\gamma$ on $\R^d\times \R^d$ such that $\gamma(A,\R^d)=\mu(A)$ and $\gamma(\R^d,A)=\nu(A)$ for all $A\subseteq\R^d$.
Further, let $c\colon\R^d\times\R^d\to\R$ be some continuous cost function which is bounded from below.
Then, the optimal transport cost $\mathrm{OT}_c(\mu,\nu)$ is defined by
\begin{equation}\label{eq_OT}
\mathrm{OT}_c(\mu,\nu)=\inf_{\gamma\in\Pi(\mu,\nu)}\int_{\R^d\times\R^d}c(x,y)d\gamma(x,y).
\end{equation}
Note, that one can show that the infimum is attained.
The semi-dual formulation of \eqref{eq_OT} is given by (see Theorem 1.42 from \cite{santambrogio2015optimal})
\begin{align}\label{eq_dualOT}
\quad\mathrm{OT}_c(\mu,\nu)=\sup_{\psi\in L^1(\nu)}\int_{\R^d}\psi^c(x) d\mu(x)+\int_{\R^d} \psi(y) d\nu(y),
\end{align}
where $\psi^c(x)=\inf_{x\in\R^d}\{c(x,y)-\psi(y)\}$ is the $c$-transform of $\psi$.
Again, one can show that the supremum is attained.

In the specific case where $\mu=\frac1N\sum_{i=1}^N\delta_{x_i}$ and $\nu=\frac1M\sum_{i=1}^M\delta_{y_i}$ are discrete measures, $\psi$ can be viewed as a vector $(\psi_i)_{i=1}^M\in\R^M$ with $\psi_i= \psi(y_i)$ and the dual formulation can be rewritten as
\begin{align*}
\mathrm{OT}_c(\mu,\nu)&=\max_{\psi\in \R^M} \Big( \frac1N\sum_{i=1}^N\psi^c(x_i)+\frac1M\sum_{i=1}^M \psi_i \Big),\\
\psi^c(x)&=\min_{i=1,...,M}\{c(x,y_i)-\psi_i\}.
\end{align*}

Finally, this optimal transport cost is related to the so so-called Wasserstein-$p$ distance when taking the cost function $c(x,y)=\|x-y\|_p^p$. The Wasserstein-$p$ distance is then defined as
\begin{equation}
W_p^p(\mu,\nu)=\mathrm{OT}_c(\mu,\nu).
\end{equation}

In this case, $W_p$ is a metric on the space of all probability measures on $\R^d$ (see proposition 5.1 from \cite{santambrogio2015optimal}). Although all results hold for a generic cost $c$, for now, we will consider the cost $c(x,y)=\|x-y\|_2^2$ and the $W_2$ distance.

\section{A Wasserstein Patch Prior}\label{sec_wasserstein_patch_prior}
\noindent
In this section, we describe the proposed superresolution method. 
First, in Subsection~\ref{sec_setup}, we introduce the Wasserstein patch prior as a regularizer
and derive our objective functional.
In Subsection~\ref{sec_minimization} we focus on the minimization of the objective functional.
Finally, in Subsection~\ref{sec_dist_diffs} we consider the case that our reference image and the ground truth have a slightly different
patch distribution and propose a way to overcome this issue.

\subsection{Proposed regularizer}\label{sec_setup}
\begin{figure*}[!t]
\centering
\scalebox{0.65}{
\begin{subfigure}[t]{0.4\textwidth}
\centering
\includegraphics[width=\textwidth]{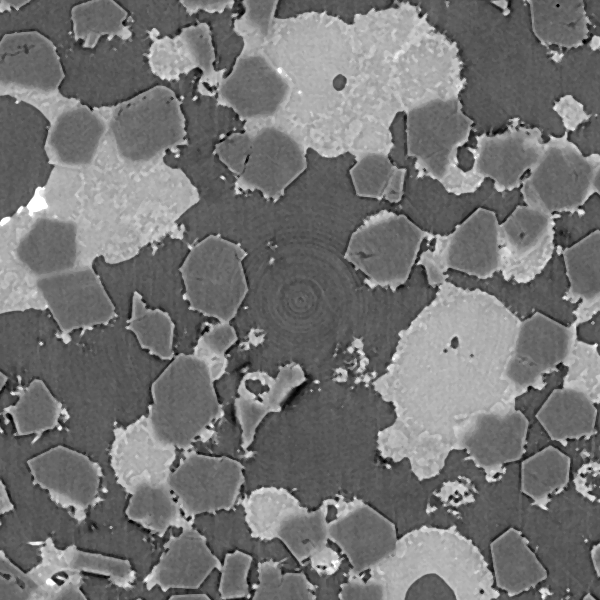}
\caption*{$x_1=x$}
\end{subfigure}
\begin{subfigure}[t]{0.2\textwidth}
\centering
\includegraphics[width=\textwidth]{imgs/img_hr}
\caption*{$x_2=Ax_1$}
\end{subfigure}
\begin{subfigure}[t]{0.1\textwidth}
\centering
\includegraphics[width=\textwidth]{imgs/img_hr}
\caption*{$x_3=A x_2$}
\end{subfigure}
\hspace{.3cm}
\begin{subfigure}[t]{0.4\textwidth}
\centering
\includegraphics[width=\textwidth]{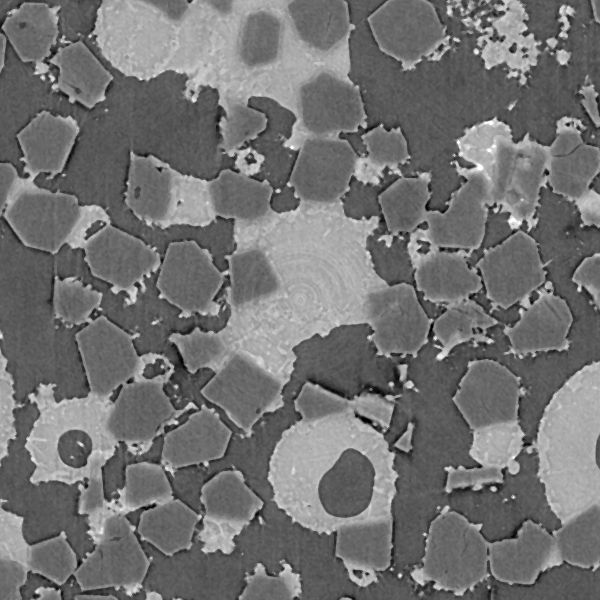}
\caption*{$\tilde x_1=\tilde x$}
\end{subfigure}
\begin{subfigure}[t]{0.2\textwidth}
\centering
\includegraphics[width=\textwidth]{imgs/img_learn}
\caption*{$\tilde x_2=A\tilde x_1$}
\end{subfigure}
\begin{subfigure}[t]{0.1\textwidth}
\centering
\includegraphics[width=\textwidth]{imgs/img_learn}
\caption*{$\tilde x_3=A \tilde x_2$}
\end{subfigure}}
\smallskip
\includegraphics[width=0.96\textwidth]{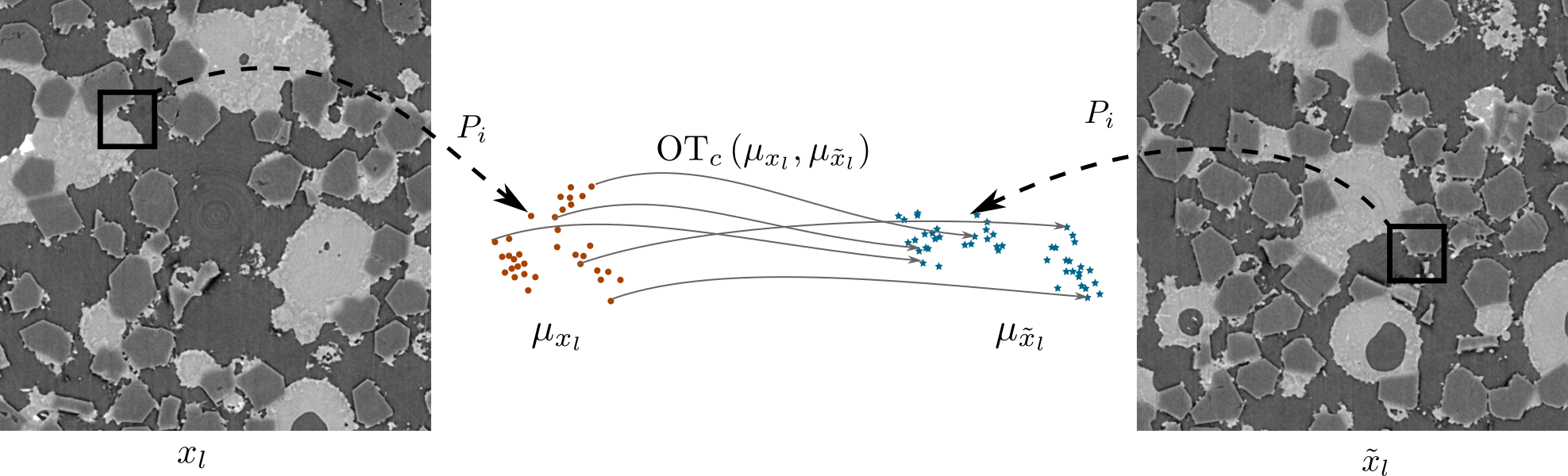}
\caption{Top: visualization of the downsampled images $x_l$ and $\tilde x_l$ for $l=1,2,3$.
Bottom: for each layer $l$, we consider the optimal transport cost between patches distributions from $x_l$ and $\tilde x_l$.}
\label{fig_multiscale}
\end{figure*}

\noindent
In this paper, we assume that we are given a reference image $\tilde x$, which has a similar patch distribution as $x$. 
This assumption is fulfilled e.g.\ when working with textures or material data, which admit lots of self-similarities.
To represent structures of different sizes within the regularizer, we consider the image $x$ at different scales.
More precisely, let $A$ be some downsampling operator. Then, we define the downsampled images $x_l$ and $\tilde x_l$ by
$x_1=x$, $x_l=A x_{l-1}$, $\tilde x_1=\tilde x$ and $\tilde x_l=A \tilde x_{l-1}$ for $l=2,...,L$.
Then, we consider the optimal transport cost between patches of $x_l$ and $\tilde x_l$ at each layer $l$. We use the regularizer
$$
R(x)=\sum_{l=1}^L \mathrm{OT}_c(\mu_{x_l},\mu_{\tilde x_l}),
$$
with $\mu_{x_l}=\frac{1}{N_l}\sum_{i=1}^{N_l} \delta_{P_i x_l}$ and
$\mu_{\tilde x_l}=\frac{1}{\tilde N_l}\sum_{i=1}^{\tilde N_l} \delta_{P_i \tilde x_l}$.
Figure~\ref{fig_multiscale} illustrates this decomposition and the considered optimal transport regularizer.

Here, the patches $(P_i x_l)_{i=1}^{N_l}$ are all patches in $x_l$, while $(P_i \tilde x_l)_{i=1}^{\tilde N_l}$ is a 
random subset of the patches in $\tilde x_l$.
Consequently, the measures $\mu_{x_l}$ and $\mu_{\tilde x_l}$ are the empirical patch distributions of
the images $x_l$ and $\tilde x_l$.
Finally, we propose to minimize the functional
$$
\mathcal J (x)=d(x,y)+\lambda R(x).
$$
Within our numerical examples, we focus on the data fidelity term 
$d(x,y)=\tfrac12\|f(x)-y\|^2$
and the cost function
$c(x,y)=\tfrac12\|x-y\|^2$
within the optimal transport term $\mathrm{OT}_c$.
In this case, $\mathcal J$ can be rewritten as
\begin{equation}\label{eq_objective}
\mathcal J (x)=\tfrac12\|f(x)-y\|^2+\lambda \sum_{l=1}^L W_2^2(\mu_{x_l},\mu_{\tilde x_l}),
\end{equation}
where $W_2^2$ is the squared Wasserstein-$2$ distance.

Note that the authors of \cite{HLPR2021} propose to find a local minimum of the regularizer $R$
to generate textures based on a reference texture $\tilde x$ for a random initialization $x$.

\subsection{Minimization of the objective function}\label{sec_minimization}
\noindent
For the minimization of $\mathcal J$, we assume that $f$ is differentiable and use a gradient descent scheme 
based on the minimization of $R$.
Here, the main challenge is to compute the derivative of $\mathrm{OT}_c\Big(\tfrac1N\sum_{i=1}^N \delta_{P_i x},\tfrac1{\tilde N}\sum_{i=1}^{\tilde N}\delta_{P_i\tilde x}\Big)$ with respect to $x$.
For this, we follow the lines of \cite{HLPR2021}.

Using the dual formulation of optimal transport, we obtain that
$$
\mathrm{OT}_c\Big(\tfrac1N\sum_{i=1}^N \delta_{P_i x},\tfrac1{\tilde N}\sum_{i=1}^{\tilde N}\delta_{P_i\tilde x}\Big)=\max_{\psi\in\R^N} F(\psi,x),$$
\begin{equation}
\label{DefF}
F(\psi,x)\coloneqq\frac1N\sum_{i=1}^N\psi^c(P_ix)+\frac1{\tilde N}\sum_{i=1}^{\tilde N}\psi_i,
\end{equation}
and $\psi^c(x)=\min_{j\in\{1,...,\tilde N\}} \{c(x,P_j \tilde x)-\psi_j\}$ is the $c$-transform of $\psi$.
Now, the following theorem was shown in \cite[Thm. 2]{HLPR2021}.

\begin{theorem}\label{thm_OT_min}
Consider $x_0$ such that $x\mapsto \mathrm{OT}_c\Big(\tfrac1N\sum_{i=1}^N \delta_{P_i x},\tfrac1{\tilde N}\sum_{i=1}^{\tilde N}\delta_{P_i\tilde x}\Big)$ and $x\mapsto F(\psi^*,x)$ 
are differentiable at $x_0$ with $\psi^*\in\argmax_{\psi} F(\psi,x_0)$.
Then it holds
\begin{equation}
\nabla_x \mathrm{OT}_c\Big(\tfrac1N\sum_{i=1}^N \delta_{P_i x_0},\tfrac1{\tilde N}\sum_{i=1}^{\tilde N}\delta_{P_i\tilde x}\Big)=\nabla_x F(\psi^*,x_0).
\end{equation}
\end{theorem}

With our choice of cost function $c(x,y)=\tfrac12\|x-y\|^2$ resulting in $\mathrm{OT}_c=\tfrac12W_2^2$,
we get
$$
\psi^c(P_ix)=\min_{j\in\{1,...,\tilde N\}} (\tfrac12\|P_i x- P_j \tilde x\|^2-\psi_j).
$$
Note that for almost every $x$ the set of minimizers
$$
\sigma_\psi(i)=\argmin_{j\in\{1,...,\tilde N\}} (\tfrac12\|P_i x- P_j \tilde x\|^2-\psi_j)
$$
is single-valued. Hence,
\begin{align*}
\nabla_x \psi^c(P_ix)&=\nabla_x(\tfrac12\|P_i x- P_{\sigma_{\psi}(i)} \tilde x\|^2)\\&=P_i^\tT P_i x - P_i^\tT P_{\sigma_\psi(i)}\tilde x.
\end{align*}

Let $\psi^*\in\argmax_\psi F(\psi,x)$. By definition of $F$, see \eqref{DefF}, and Theorem~\ref{thm_OT_min}, the gradient $\nabla_x W_2^2(\mu_x,\mu_{\tilde x})$ for almost every $x$ is given by
\begin{align*}
\nabla_x W_2^2(\mu_x,\mu_{\tilde x}) &=\nabla_x F(\psi^*,x)\\&=\frac1N\sum_{i=1}^NP_i^\tT P_i x - P_i^\tT P_{\sigma_{\psi^*}(i)}\tilde x.
\end{align*}
Note that for computing this gradient it is required to compute $\psi^*$. For doing so,
the authors of \cite{HLPR2021} suggest to use a (stochastic) gradient ascent as proposed in \cite{GCPB2016}.

Summarized, the gradient of $\mathcal J(x)$ in \eqref{eq_objective} can be computed by Algorithm~\ref{alg_gradient_J}.

\begin{algorithm*}[!t]
\caption{Gradient computation}\label{alg_gradient_J}
\begin{algorithmic}
\For{$l=1,...,L$}
\State Use stochastic gradient ascent to compute
\begin{align*}
\psi^l\in\argmax_{\psi}F_l(\psi,x_l)\coloneqq\frac{1}{N_l}\sum_{i=1}^{N_l}\psi^c(P_ix_l)
+\frac1{\tilde N_l}\sum_{j=1}^{\tilde N_l}\psi_j.
\end{align*}
\EndFor
\State Compute $\nabla\mathcal J(x)$ as
\begin{align*}
\nabla\mathcal J(x)=\nabla f(x) (f(x)-y)+\lambda \sum_{l=1}^L (A^l)^\tT \Big(\frac1{N_l}\sum_{i=1}^{N_l}P_i^\tT P_i x_l-P_i^\tT P_{\sigma_{\psi^l}(i)}\tilde x_l\Big),
\end{align*}
where $\nabla f(x)$ is the Jacobian of $f$ at $x$.
\end{algorithmic}
\end{algorithm*}

\subsection{Compensation of slightly different patch distributions}\label{sec_dist_diffs}
\noindent
So far, we assumed that the patch distributions in the (unknown) ground truth $x$ and the reference image $\tilde x$ are exactly equal. 
This assumption is not realistic in practice. To compensate for slight differences of $\mu_x$ and 
$\mu_{\tilde x}$, we introduce the new operator $g\colon\R^{m+2p\times n+2p}\to\R^d$ defined by
$g(x)=f(Cx)$, where $C\colon \R^{m+2p\times n+2p}\to\R^{m\times n}$ crops the middle $m\times n$ pixels from $x\in\R^{m+2p\times n+2p}$.
Then, we minimize
$$
\mathcal I(x)=\tfrac12\|g(x)-y\|^2+\lambda \sum_{l=1}^L W_2^2(\mu_{x_l},\mu_{\tilde x_l}).
$$
Finally, our reconstruction is given by
$$
\hat x=Cz,\quad\text{with}\quad z\in\argmin_{x} \mathcal I(x).
$$

Note that the data fidelity term $\tfrac12\|g(x)-y\|^2$ is by definition not affected by the boundary
of size $p$ in $x$.
On the other side, the boundary of $x$ influences the patch distribution of $x$ such that it can compensate small 
differences between the patch distributions of the reference image $\tilde x$ and the unknown ground truth 
$x$.

\section{Numerical Results for Superresolution}\label{sec_numerics}
\noindent
In this section, we apply the Wasserstein patch regularization for the problem of superresolution of two- and three-dimensional images. In this case, the forward operator $f$ usually is a composition of
a blur operator and a downsampling operator. 
We demonstrate the performance of our approach by using images of materials' microstructures obtained by synchrotron micro-computed tomography (s$\mu$CT). Additionally, we consider synthetic images that were obtained by simulating the process of a serial sectioning imaging technique combining focused ion beam milling with imaging by a scanning electron microscope (FIB-SEM). Finally, the approach is applied to a real FIB-SEM image stack. 

\paragraph{Evaluation of results} For evaluating the quality of our results we use three different error measures.
First, we use the peak-signal to noise ration (PSNR). For two images $x$ and $y$ on $[0,1]^{m\times n}$ it is defined as
$$
\mathrm{PSNR}(x,y)=-10\log_{10}(\tfrac{1}{mn}\|x-y\|^2).
$$
Second, we measure the sharpness of our results by using the so-called blur effect \cite{CDLN2007}. This metric is based on comparing an input image $x$ with a blurred version $x_{\blur}$. For sharp images $x$, the difference should be very pronounced while it will be small for blurred $x$.
The blur effect is normalized to $[0,1]$, where a small blur effect indicates that $x$ is very sharp while a large blur effect means that $x$ is very blurry.
Finally, we use the learned perceptual image patch similarity (LPIPS) \cite{ZIESW2018}\footnote{We use the implementation \url{https://github.com/richzhang/PerceptualSimilarity}, version 0.1.} for measuring the perceptual similarity of our results and the ground truth.
The basic idea of LPIPS is to compare the feature maps extracted from some deep neural network that is trained for some classical imaging task which is not necessarily related to our original problem.
A small value of LPIPS indicates a high perceptual similarity.
Note that LPIPS was originally proposed and implemented for 2D images. 
Although LPIPS could probably be extended to 3D images 
(see \cite{SGST2020} for some work in this direction), this is not within the scope of our paper.
For the 3D data, we will use the mean of the LPIPS values over all slices of the image.

All numerical examples are implemented in Python and PyTorch and are based on the code of \cite{HLPR2021}\footnote{available at \url{https://github.com/ahoudard/wgenpatex}}.

\paragraph{Comparison to established methods} 
We compare our results with some established methods.

First, we compute the bi- and tricubic interpolation \cite{K1981,LM2005}. These interpolations are based on the local approximation of the image by polynomials of degree $3$.

Second, we compare our results with the $L^2$-TV reconstruction. It was proposed in \cite{ROF1992} and is defined as a solution $\hat x$ of
\begin{equation}\label{eq_l2tv}
\hat x\in\argmin_{x}\frac12\|f(x)-y\|^2+\lambda \mathrm{TV}(x),
\end{equation}
where $\mathrm{TV}$ is the total variation. It is defined by $\mathrm{TV}(x)=\|D x\|_1$, where $D$ is the matrix which maps $x$ onto the vector $Dx$ which contains all differences of neighboured pixels.
The scalar $\lambda>0$ serves as weighting between data fidelity term and regularizer.
Note that the objective \eqref{eq_l2tv} is convex, such that the solution can be easily computed by classical methods like the alternating dirction method of multipliers \cite{G1983,GM1976} or a Chambolle-Pock algorithm \cite{CP2011}.

For the 2D images, we additionally consider the expected patch log likelihood (EPLL) prior, as proposed by Zoran and Weiss in \cite{ZW2011} and
further investigated in \cite{PDDN2019}\footnote{We do not reimplement the EPLL algorithm. 
Instead we use the code of \cite{PDDN2019} available at \url{https://github.com/pshibby/fepll_public}.}.
The idea of EPLL is to assume that the patch distribution can be approximated by a Gaussian mixture model which is fitted to the patch distribution of the reference image. Once the GMM $p$ is estimated, the minimization problem
$$
\hat x\in\argmin_x \frac{1}{2}\|f(x)-y\|^2-\lambda\sum_{i=1}^N \log p(P_ix)
$$ 
is solved approximately.
For denoising tasks, EPLL has shown great performance and beats several classical methods as e.g.\ BM3D \cite{DKE2012}, see \cite{ZW2011}.

Finally, we compare our 2D results with two neural network based approaches, namely the deep image prior (DIP) \cite{UVL2018} and a Plug-and-Play forward backward splitting (PnP-FBS) \cite{SVWB2016, VBW2013}.
The idea of DIP \cite{UVL2018}\footnote{We use the original implementation from \cite{UVL2018} available at \url{https://github.com/DmitryUlyanov/deep-image-prior}} is to solve the optimization problem
$$
\hat \theta\in\argmin_{\theta}\|f(G_\theta(z))-y\|^2,
$$
where $G_\theta$ is a convolutional neural network with parameters $\theta$ and $z$ is a randomly chosen input. 
Then, the reconstruction $\hat x$ is given by $\hat x=G_\theta(z)$.
It was shown in \cite{UVL2018}, that DIP admits competitive results for many inverse problems.

Plug and Play methods were first introduced in \cite{VBW2013}. The main idea is to consider a classical algorithm from convex optimization
and replace the proximal operator with respect to the regularizer by a more general denoiser.
More precisely, we modify the forward backward splitting algorithm for minimizing the function $F(x)=d(x)+R(x)$ with $d(x)=\tfrac12\|f(x)-y\|^2$ given by
\begin{equation}\label{eq_FBS}
x_{r+1}=\mathrm{prox}_{\eta R}(x_r-\eta\nabla d(x_r))
\end{equation}
by the iteration
\begin{equation}\label{eq_PnP_FBS}
x_{r+1}= \mathcal D(x_r-\eta\nabla d(x_r)),
\end{equation}
where $\mathcal D$ is a neural network trained for denoising natural images. 
Here, we use the DRUNet from \cite{ZLZZ2021} as denoiser $\mathcal D$ and run the iteration \eqref{eq_PnP_FBS} for $100$ iterations. Then, we
set the reconstruction to be $\hat x=x_{100}$.
It was shown in \cite{CWE2016,GJNMU2018,HNS2021,MMHC2017,O2017,ZLZZ2021} that Plug-and-Play methods can achieve state-of-the-art performance for several applications.

\begin{remark}
To achieve computational efficiency, we do not use all patches within the measure $\mu_{\tilde x}$ arising from the reference image but only a subset of $4000$ patches.
For the measure $\mu_x$ arising from our reconstruction, we use all patches.
\end{remark}

\subsection{Synchrotron computed tomography data}\label{sec_material_data}
\begin{figure}
\begin{subfigure}[t]{0.245\textwidth}
\includegraphics[width=\textwidth]{imgs/img_hr}
\caption*{Ground truth SiC}
\end{subfigure}\hfill
\begin{subfigure}[t]{0.245\textwidth}
\includegraphics[width=\textwidth]{imgs/img_learn}
\caption*{Reference SiC}
\end{subfigure}\hfill
\begin{subfigure}[t]{0.245\textwidth}
\includegraphics[width=\textwidth]{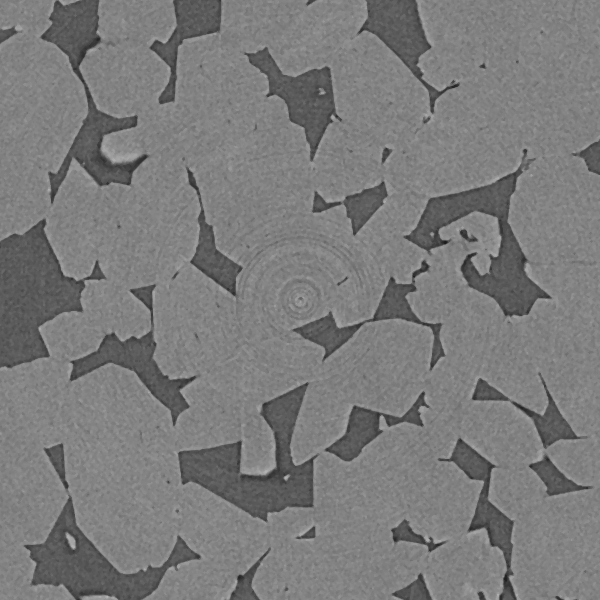}
\caption*{Ground truth FS}
\end{subfigure}\hfill
\begin{subfigure}[t]{0.245\textwidth}
\includegraphics[width=\textwidth]{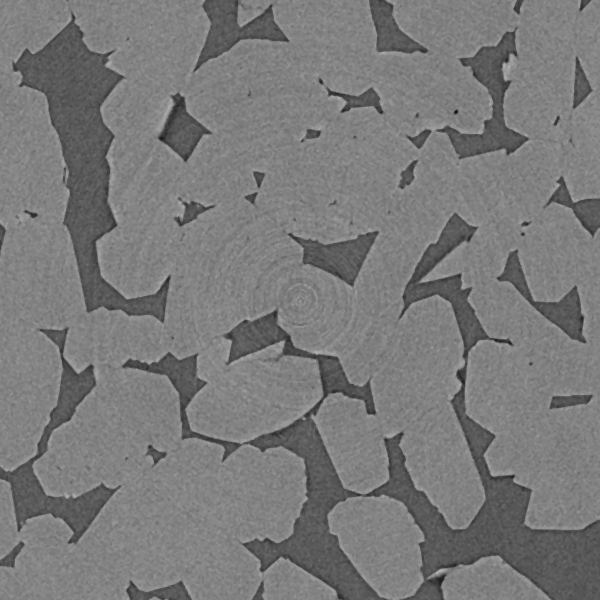}
\caption*{Reference FS}
\end{subfigure}
\caption{Ground truth and reference images for the two different materials
in our numerical examples. 
Left: SiC Diamonds, 
right: Fontainebleau Sandstone (FS).}
\label{fig_ground_truth}
\end{figure}

\noindent
First, we consider material data which was also used in \cite{HNABBSS2020}.
Here, a series of multi-scale 3D images has been acquired by synchrotron micro-computed tomography at the SLS beamline TOMCAT. Samples of two materials were selected to provide 3D images having different levels of complexity:
\begin{itemize}
\item[-] The first one is a composite ("SiC Diamonds") obtained by microwave sintering of silicon and diamonds, see \cite{vaucher2007line}.
\item[-] The second one is a sample of Fontainebleau sandstone ("FS"), a rather homogeneous natural rock that is commonly used in the oil industry for flow experiments.
\end{itemize}
The ground truth and the reference images in our two-dimensional experiments are given in Figure~\ref{fig_ground_truth} and have the size $600\times 600$ pixels with a pixel spacing of $1.625$ {\textmu}m.
Since we require that the forward operator $f$ is known, we use the real material images as high resolution ground-truth and as reference image. 
To obtain the corresponding low resolution image we artificially downsample our high resolution ground-truth using a known predefined forward operator $f$.

For this example, we choose $f$ to be a convolution with a $16\times 16$ Gaussian blur kernel with standard deviation $2$ and stride $4$ (without padding).
After applying $f$, we add Gaussian noise $\eta\sim\mathcal N(0,0.01^2)$.
Thus, the low resolution observation is an image of size $147\times147$.
For the downsampling operator $A$ within the definition of the regularizer $R$, we choose a convolution with a $4\times 4$ Gaussian
blur kernel with standard deviation $1$ and stride $2$ (without padding). Note that this downsampling operator was also used in \cite{HLPR2021}.
Further, we set the number of scales to $L=2$.
Since we cannot expect a useful reconstruction close to the boundary, we crop the middle $520\times 520$ pixels from the reconstruction
for evaluating the error measures.

Further, we add a boundary of size $p=20$ as described in Section~\ref{sec_dist_diffs}.
As initialization, we use the bicubic interpolation. 
The bicubic interpolation with magnification factor $4$ of a $147\times 147$ image has size $588\times588$, 
but our initialization has size $640\times 640$. Hence, we set the boundary of our initialization to random noise 
uniformly distributed on $[0,1]$.\\

\begin{figure}
\centering
\begin{subfigure}[t]{0.2\textwidth}
\includegraphics[width=\textwidth]{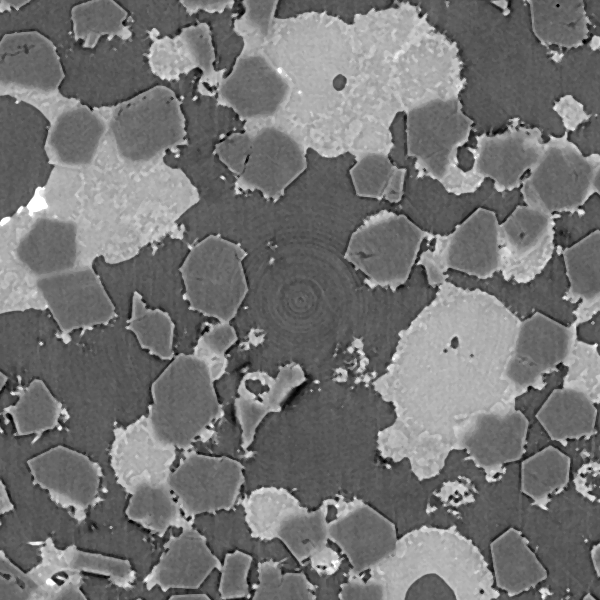}
\end{subfigure}
\begin{subfigure}[t]{0.2\textwidth}
\includegraphics[width=\textwidth]{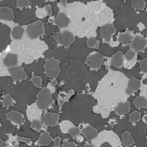}
\end{subfigure}
\begin{subfigure}[t]{0.2\textwidth}
\includegraphics[width=\textwidth]{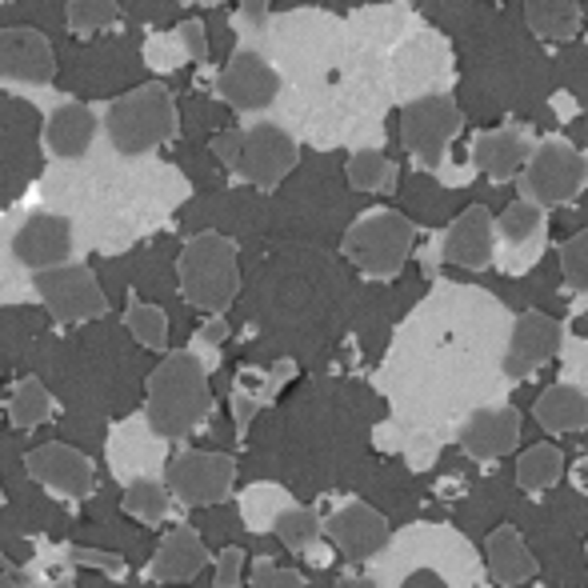}
\end{subfigure}
\begin{subfigure}[t]{0.2\textwidth}
\includegraphics[width=\textwidth]{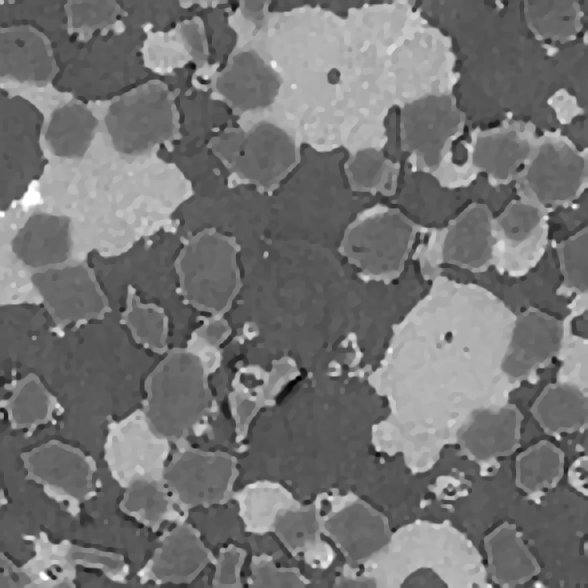}
\end{subfigure}

\begin{subfigure}[t]{0.2\textwidth}
\includegraphics[width=\textwidth]{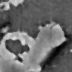}
\end{subfigure}
\begin{subfigure}[t]{0.2\textwidth}
\includegraphics[width=\textwidth]{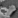}
\end{subfigure}
\begin{subfigure}[t]{0.2\textwidth}
\includegraphics[width=\textwidth]{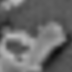}
\end{subfigure}
\begin{subfigure}[t]{0.2\textwidth}
\includegraphics[width=\textwidth]{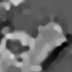}
\end{subfigure}

\begin{subfigure}[t]{0.2\textwidth}
\includegraphics[width=\textwidth]{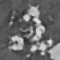}
\caption*{HR image}
\end{subfigure}
\begin{subfigure}[t]{0.2\textwidth}
\includegraphics[width=\textwidth]{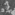}
\caption*{LR image}
\end{subfigure}
\begin{subfigure}[t]{0.2\textwidth}
\includegraphics[width=\textwidth]{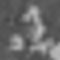}
\caption*{bicubic}
\end{subfigure}
\begin{subfigure}[t]{0.2\textwidth}
\includegraphics[width=\textwidth]{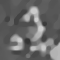}
\caption*{$L^2$-TV}
\end{subfigure}

\begin{subfigure}[t]{0.2\textwidth}
\includegraphics[width=\textwidth]{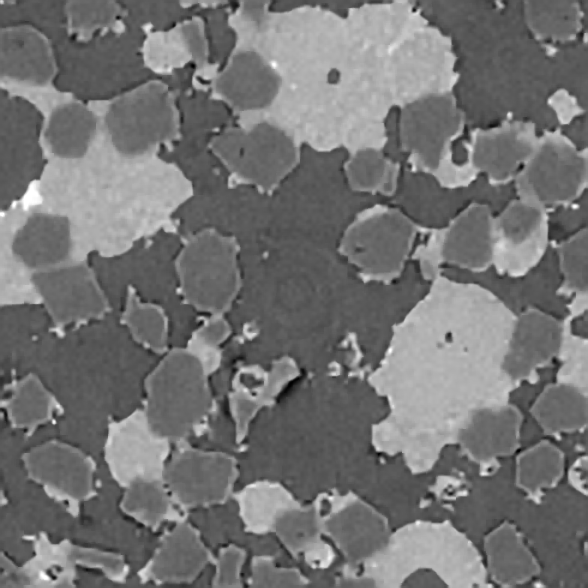}
\end{subfigure}
\begin{subfigure}[t]{0.2\textwidth}
\includegraphics[width=\textwidth]{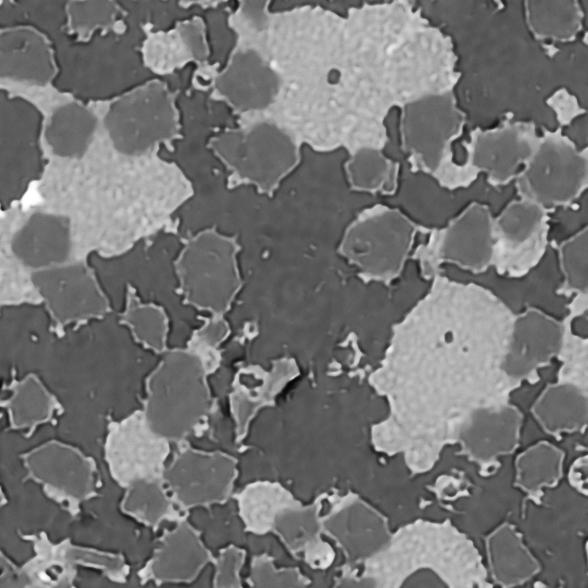}
\end{subfigure}
\begin{subfigure}[t]{0.2\textwidth}
\includegraphics[width=\textwidth]{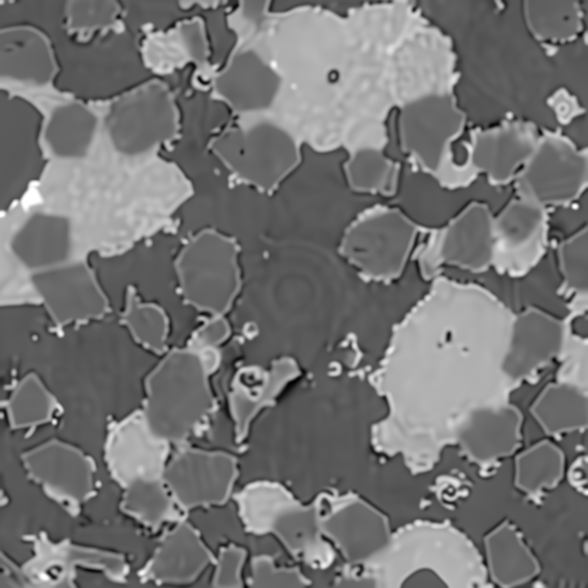}
\end{subfigure}
\begin{subfigure}[t]{0.2\textwidth}
\includegraphics[width=\textwidth]{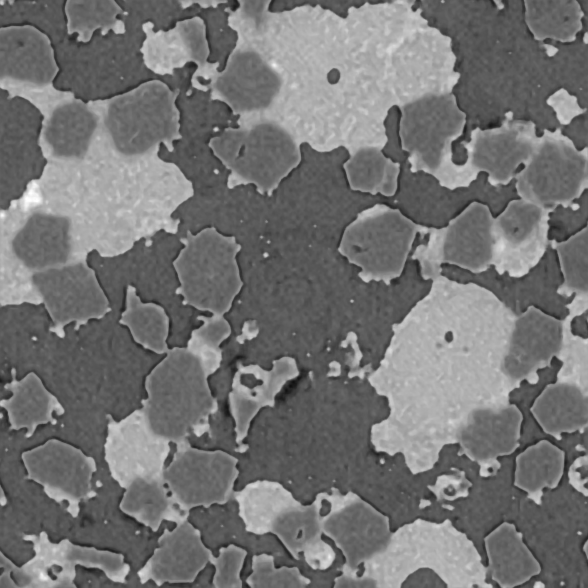}
\end{subfigure}

\begin{subfigure}[t]{0.2\textwidth}
\includegraphics[width=\textwidth]{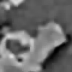}
\end{subfigure}
\begin{subfigure}[t]{0.2\textwidth}
\includegraphics[width=\textwidth]{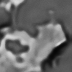}
\end{subfigure}
\begin{subfigure}[t]{0.2\textwidth}
\includegraphics[width=\textwidth]{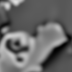}
\end{subfigure}
\begin{subfigure}[t]{0.2\textwidth}
\includegraphics[width=\textwidth]{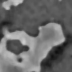}
\end{subfigure}

\begin{subfigure}[t]{0.2\textwidth}
\includegraphics[width=\textwidth]{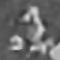}
\caption*{EPLL}
\end{subfigure}
\begin{subfigure}[t]{0.2\textwidth}
\includegraphics[width=\textwidth]{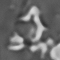}
\caption*{DIP}
\end{subfigure}
\begin{subfigure}[t]{0.2\textwidth}
\includegraphics[width=\textwidth]{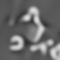}
\caption*{PnP-FBS}
\end{subfigure}
\begin{subfigure}[t]{0.2\textwidth}
\includegraphics[width=\textwidth]{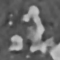}
\caption*{$W_2^2$-regularized}
\end{subfigure}
\caption{Reconstruction of the high resolution image "SiC Diamonds" using different methods.  Top: Full image, middle and bottom:
zoomed-in parts.}
\label{fig_results_diam}
\end{figure}

\begin{figure}
\centering
\begin{subfigure}[t]{0.2\textwidth}
\includegraphics[width=\textwidth]{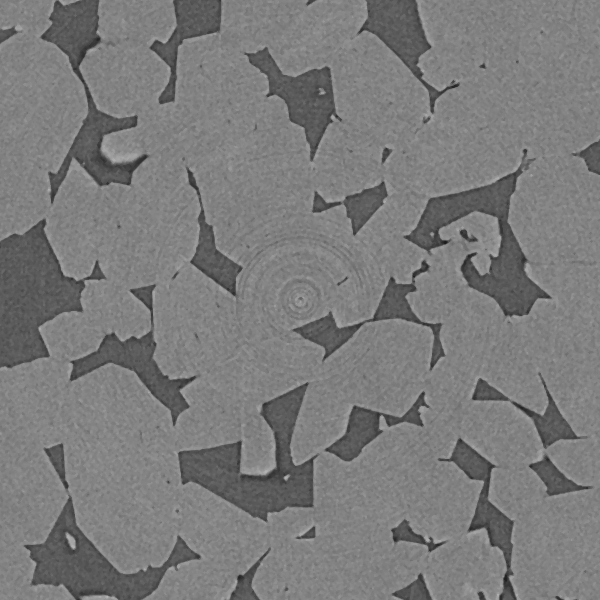}
\end{subfigure}
\begin{subfigure}[t]{0.2\textwidth}
\includegraphics[width=\textwidth]{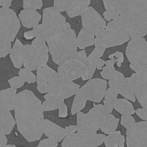}
\end{subfigure}
\begin{subfigure}[t]{0.2\textwidth}
\includegraphics[width=\textwidth]{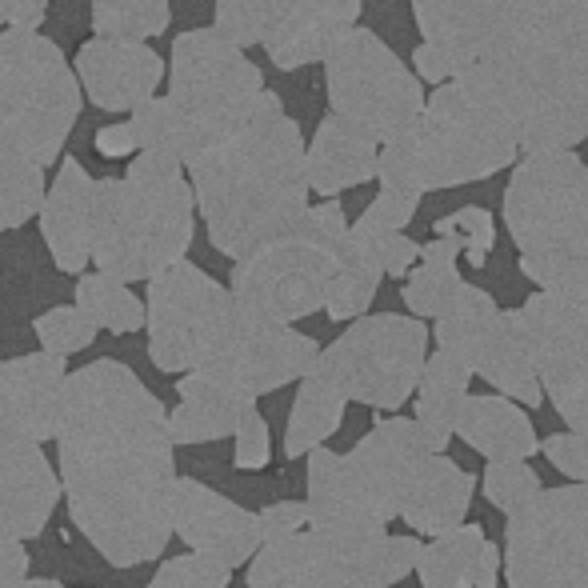}
\end{subfigure}
\begin{subfigure}[t]{0.2\textwidth}
\includegraphics[width=\textwidth]{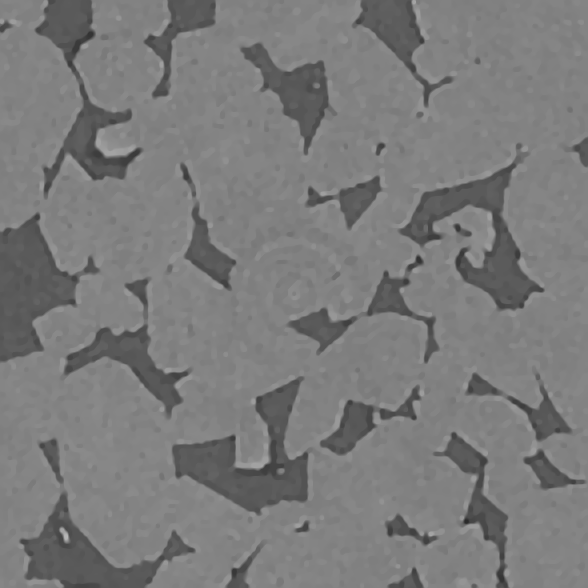}
\end{subfigure}

\begin{subfigure}[t]{0.2\textwidth}
\includegraphics[width=\textwidth]{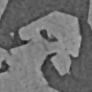}
\end{subfigure}
\begin{subfigure}[t]{0.2\textwidth}
\includegraphics[width=\textwidth]{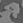}
\end{subfigure}
\begin{subfigure}[t]{0.2\textwidth}
\includegraphics[width=\textwidth]{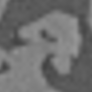}
\end{subfigure}
\begin{subfigure}[t]{0.2\textwidth}
\includegraphics[width=\textwidth]{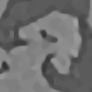}
\end{subfigure}

\begin{subfigure}[t]{0.2\textwidth}
\includegraphics[width=\textwidth]{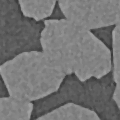}
\caption*{HR image}
\end{subfigure}
\begin{subfigure}[t]{0.2\textwidth}
\includegraphics[width=\textwidth]{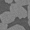}
\caption*{LR image}
\end{subfigure}
\begin{subfigure}[t]{0.2\textwidth}
\includegraphics[width=\textwidth]{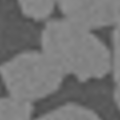}
\caption*{bicubic}
\end{subfigure}
\begin{subfigure}[t]{0.2\textwidth}
\includegraphics[width=\textwidth]{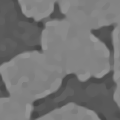}
\caption*{$L^2$-TV}
\end{subfigure}

\begin{subfigure}[t]{0.2\textwidth}
\includegraphics[width=\textwidth]{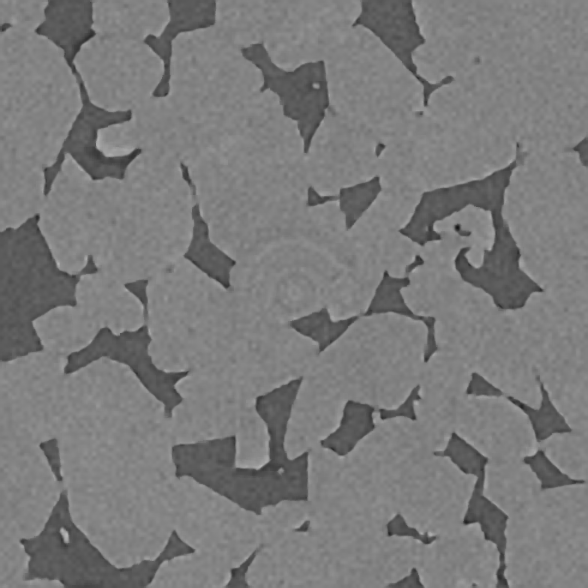}
\end{subfigure}
\begin{subfigure}[t]{0.2\textwidth}
\includegraphics[width=\textwidth]{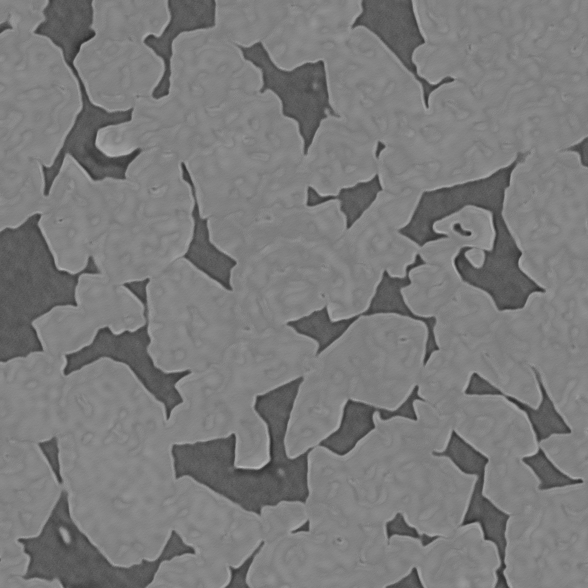}
\end{subfigure}
\begin{subfigure}[t]{0.2\textwidth}
\includegraphics[width=\textwidth]{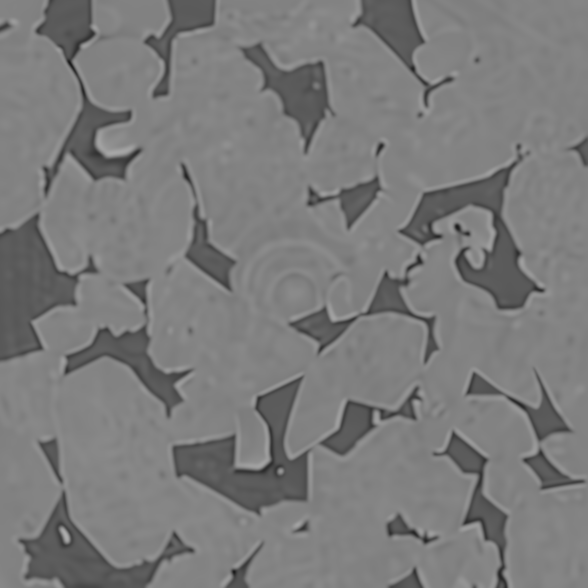}
\end{subfigure}
\begin{subfigure}[t]{0.2\textwidth}
\includegraphics[width=\textwidth]{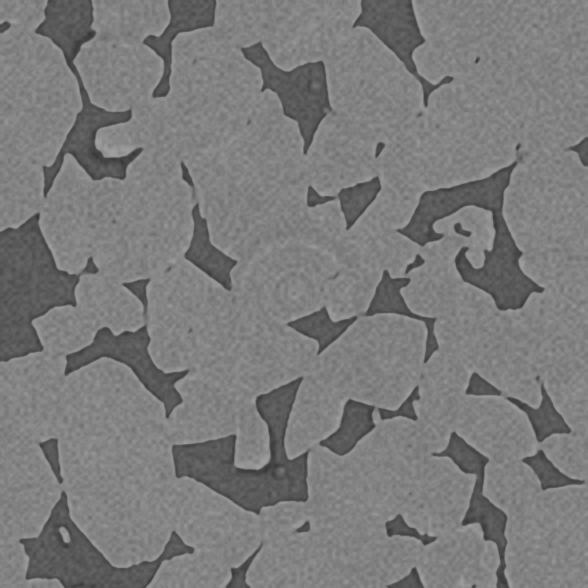}
\end{subfigure}

\begin{subfigure}[t]{0.2\textwidth}
\includegraphics[width=\textwidth]{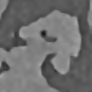}
\end{subfigure}
\begin{subfigure}[t]{0.2\textwidth}
\includegraphics[width=\textwidth]{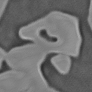}
\end{subfigure}
\begin{subfigure}[t]{0.2\textwidth}
\includegraphics[width=\textwidth]{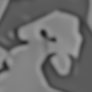}
\end{subfigure}
\begin{subfigure}[t]{0.2\textwidth}
\includegraphics[width=\textwidth]{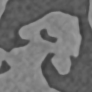}
\end{subfigure}

\begin{subfigure}[t]{0.2\textwidth}
\includegraphics[width=\textwidth]{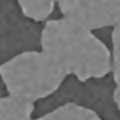}
\caption*{EPLL}
\end{subfigure}
\begin{subfigure}[t]{0.2\textwidth}
\includegraphics[width=\textwidth]{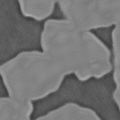}
\caption*{DIP}
\end{subfigure}
\begin{subfigure}[t]{0.2\textwidth}
\includegraphics[width=\textwidth]{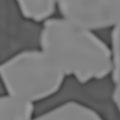}
\caption*{PnP-FBS}
\end{subfigure}
\begin{subfigure}[t]{0.2\textwidth}
\includegraphics[width=\textwidth]{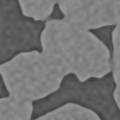}
\caption*{$W_2^2$-regularized}
\end{subfigure}
\caption{Reconstruction of the high resolution image "Fontainebleau Sandstone" using different methods. 
 Top: Full image, middle and bottom:
zoomed-in parts.}
\label{fig_results_FS}
\end{figure}

The resulting error measures are given in Table~\ref{tab_results}.
The reconstructions are shown in Figures~\ref{fig_results_diam} and \ref{fig_results_FS}.
We observe that the Wasserstein-$2$ patch prior outperforms the other methods visually and in terms of the error measures.
Even though the PSNR values of the $L^2$-TV reconstruction are better than those of the EPLL, the results of EPLL are visually better than the results of $L^2$-TV.
This can also be seen by the other error measures.

\begin{table*}
\begin{center}
\begin{tabular}{c|c|ccccccc}
&&Bicubic&$L^2$-TV&EPLL&DIP&PnP-FBS&$W_2^2$-regularized\\\hline
\multirow{3}{*}{SiC}&PSNR&$25.06$&$27.39$&$27.26$&$27.28$&$27.46$&$\mathbf{27.50}$\\
&Blur Effect&$0.5539$&$0.4038$&$0.4285$&$0.3840$&$0.4433$&$\mathbf{0.3754}$\\
&LPIPS&$0.4147$&$0.2036$&$0.2183$&$0.1964$&$0.3161$&$\mathbf{0.1612}$\\\hline
\multirow{3}{*}{FS}&PSNR&$29.04$&$31.03$&$31.12$&$\mathbf{31.20}$&$31.07$&$31.10$\\
&Blur effect&$0.4971$&$0.4073$&$0.3761$&$0.3706$&$0.4843$&$\mathbf{0.3246}$\\
&LPIPS&$0.3561$&$0.2571$&$0.1987$&$0.2185$&$0.3512$&$\mathbf{0.1515}$
\end{tabular}
\end{center}
\caption{PSNR, blur effect and LPIPS value of the high-resolution reconstruction of the 2D-material data using different methods.}
\label{tab_results}
\end{table*}

\paragraph{3D Images.}

We redo the experiments with 3D images of size $176\times176\times176$. The forward operator is the same as for the 2D images, which results in low resolution observations of size $41\times41\times41$.
The resulting error measures for the 3D images FS and SiC Diamonds are given in Table~\ref{tab_3D_results}.
Similar to the two dimensional case, the $W_2^2$-regularized reconstruction is significantly better than $L^2$-TV and the
tricubic interpolation.

\begin{table}
\begin{center}
\begin{tabular}{c|c|ccccc}
&&Tricubic&$L^2$-TV&$W_2^2$-regularized\\\hline
\multirow{3}{*}{SiC}&PSNR&$23.06$&$24.96$&$\mathbf{25.15}$\\
&Blur effect&$0.6156$&$0.4756$&$\mathbf{0.3840}$\\
&LPIPS&$0.5657$&$0.3156$&$\mathbf{0.2260}$\\\hline
\multirow{3}{*}{FS}&PSNR&$27.16$&$28.84$&$\mathbf{28.98}$\\
&Blur effect&$0.5645$&$0.4649$&$\mathbf{0.3452}$\\
&LPIPS&$0.5427$&$0.3468$&$\mathbf{0.2160}$
\end{tabular}
\end{center}
\caption{PSNR, blur effect and LPIPS value of the high-resolution reconstruction of the 3D-material data using different methods.}
\label{tab_3D_results}
\end{table}

\subsection{Synthetic FIB-SEM images}\label{sec_fib_rem}
\noindent
Second, we consider synthetic image data. They are obtained by simulating the process of focused ion beam scanning electron microscopy imaging (FIB-SEM) by using the protocol described in \cite{prill12:scanning}. The imaged geometry is a realization of a Boolean model of spheres of identical radius \cite{chiu13}. Images at several resolutions ranging from (cubical) voxel sizes of 3 nm to 18 nm were generated. For more details on the image data we refer to \cite{ROLDAN2021113291}.  As ground truth image, we will use the images with 6 nm voxel size. The low resolution images are given by the versions with spacings 9 nm, 12 nm, 15 nm and 18 nm. To reduce the noise on the images, we preprocess each image by a $3\times 3\times 3$ median filter.  Images at the different resolutions are registered, but the forward operator $f$ is unknown and its formulation is not straightforward. 
Therefore, we propose to estimate $f$ in a first step based on one registered pair of a high-resolution and a low-resolution image. Afterwards, we minimize the functional \eqref{eq_objective} using the estimation of $f$.

\paragraph{Approximation of the Forward Operator.}

In the previous examples, we used a strided convolution as downsampling operator because of its simplicity. 
However, as we now consider also non-integral magnification factors, this appears to be infeasible. 
Thus, we assume that our forward operator is given by $\hat f(x)=S(k*x+b)$ for a $15\times 15\times 15$ blur kernel $k$, a bias $b\in\R$ and a downsampling operator $S$.
To ensure that $\hat f$ describes our data as good as possible, we will adapt the blur kernel $k$ and the bias $b$ to the registered pair $(\tilde x,\tilde y)$ as described in the paragraph below.

Further, for the downsampling operator $S$, we make use of Fourier transforms.
Given a 3D-image $x\in\R^{n_x,n_y,n_z}$ the three-dimensional discrete Fourier transform (DFT)
is defined by $\mathcal F_{n_x,n_y,n_z}\coloneqq \mathcal F_{n_x}\otimes \mathcal F_{n_y}\otimes \mathcal F_{n_z}$, 
where $\mathcal F_n=(\exp(-2\pi i k l/n))_{k,l=0}^{n-1}$. Now, the downsampling operator $S\colon\R^{m_x,m_y,m_z}\to\R^{n_x,n_y,n_z}$ 
is given by
$$
S=\frac{n_xn_yn_z}{m_xm_ym_z}\mathcal F_{n_x,n_y,n_z}^{-1} D \mathcal F_{m_x,m_y,m_z},
$$
where for $x\in\C^{m_x,m_y,m_z}$ the $(i,j,k)$-th entry of $D(x)$ is given by
$x_{i',j',k'}$, where
\begin{align}
i'=\begin{cases}i, &$if $i\leq\frac{n_x}{2},\\i+m_x-n_x, &$otherwise.$\end{cases}
\end{align}
and $j'$ and $k'$ are defined analogously. 
Thus, the operator $S$ generates a downsampled version $S(x)$ of an image $x$ by removing the high-frequency part from $x$.
Note that even if the Fourier matrix $\mathcal F_{n_x,n_y,n_z}$ is complex valued, the range of $S$ is real-valued, as $D$ preserves Hermitian-symmetric spectra.

\paragraph{Estimation of Blur Kernel and Bias.}

We assume that we have given images $\tilde x\in\R^{m_x,m_y,m_z}$ and $\tilde y\in\R^{n_x,n_y,n_z}$ related by $\tilde y\approx S(k*\tilde x+b)$, where the blur kernel $k\in\R^{15\times15\times15}$ and the bias $b\in\R$ are unknown.
In the following, we aim to reconstruct $k$ and $b$ from $\tilde x$ and $\tilde y$. 
Here, we use the notations $N=n_xn_yn_z$ and $M=m_xm_ym_z$. Further let $\tilde k\in\R^{m_x,m_y,m_z}$ be the kernel $k$ padded with zeros such that it still corresponds to the same convolution as $k$, but has size $m_x\times m_y\times m_z$.

Applying the DFT on both sides of $y=S(k*\tilde x+b)=S(\tilde k*\tilde x+b)$ and using the definition of $S$, we obtain that
$$
\hat y =\frac{N}{M}D(\hat k\odot \hat x + M b e)=\frac{N}{M}D(\hat k)\odot D(\hat x)+Nbe,
$$
where $\hat y=\mathcal F_{n_x,n_y,n_z}\tilde y$, $\hat x=\mathcal F_{m_x,m_y,m_z}\tilde x$, $\hat k=\mathcal F_{m_x,m_y,m_z}\tilde k$, $\odot$ is the elementwise product and $e$ denotes the first unit vector (i.e.\ $e_{0,0,0}=1$ and all other entries are zero). Now, we can conclude that
$$
D(\hat k)=\frac{M}{N}\hat y\oslash D(\hat x)-  \frac{M b}{\hat x_{0,0,0}} e,
$$
where $\oslash$ is the elementwise quotient. In practice, we stabilize this quotient by increasing the absolute value of $D(\hat x)$ by $10^{-5}$ while retaining the phase.
Thus, assuming that the high-frequency part of $k$ is negligible (i.e., that $D^T D k=k$), we can approximate $\hat k$ by
$$
\hat k\approx \frac{M}{N} D^\tT \hat y \oslash D(\hat x) -  \frac{Mb}{\hat x_{0,0,0}} e.
$$
Applying the inverse DFT this becomes
$$
\tilde k\approx \mathcal F_{m_x,m_y,m_z}^{-1}\Big(\frac{M}{N} D^\tT (\hat y \oslash D(\hat x))\Big) - \frac{ b}{\hat x_{0,0,0}}.
$$
Using the assumption that $\tilde k$ is zero outside of the $15\times15\times15$ patch, where $k$ is located, we can estimate $b$ by taking the mean over all pixels of $\mathcal F_{m_x,m_y,m_z}^{-1}\Big(\frac{M}{N} D^\tT (\hat y \oslash D(\hat x))\Big)$ outside of this $15\times15\times15$ patch.
Afterwards, we estimate $k$ by reprojecting
$$
\mathcal F_{m_x,m_y,m_z}^{-1}\Big(\frac{M}{N} D^\tT (\hat y \oslash D(\hat x))\Big) - \frac{ b}{\hat x_{0,0,0}}
$$
to the set of all real $15\times15\times15$ kernels.

\paragraph{Reconstruction.}

\begin{figure*}[!t]
\centering
\begin{subfigure}[t]{0.19\textwidth}
\includegraphics[width=\textwidth]{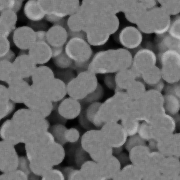}
\end{subfigure}
\begin{subfigure}[t]{0.19\textwidth}
\includegraphics[width=\textwidth]{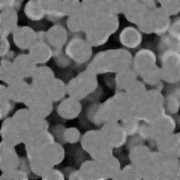}
\end{subfigure}
\begin{subfigure}[t]{0.19\textwidth}
\includegraphics[width=\textwidth]{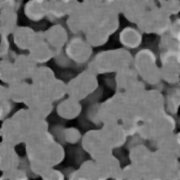}
\end{subfigure}
\begin{subfigure}[t]{0.19\textwidth}
\includegraphics[width=\textwidth]{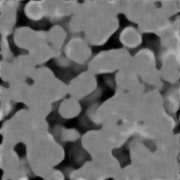}
\end{subfigure}
\begin{subfigure}[t]{0.19\textwidth}
\includegraphics[width=\textwidth]{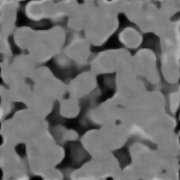}
\end{subfigure}
\begin{subfigure}[t]{0.19\textwidth}
\includegraphics[width=\textwidth]{imgs/FIB/6_xy_90}
\caption*{Ground truth xy}
\end{subfigure}
\begin{subfigure}[t]{0.19\textwidth}
\includegraphics[width=\textwidth]{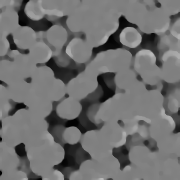}
\caption*{9 xy}
\end{subfigure}
\begin{subfigure}[t]{0.19\textwidth}
\includegraphics[width=\textwidth]{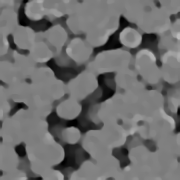}
\caption*{12 xy}
\end{subfigure}
\begin{subfigure}[t]{0.19\textwidth}
\includegraphics[width=\textwidth]{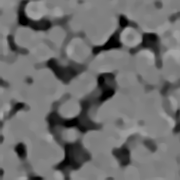}
\caption*{15 xy}
\end{subfigure}
\begin{subfigure}[t]{0.19\textwidth}
\includegraphics[width=\textwidth]{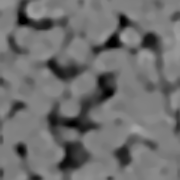}
\caption*{18 xy}
\end{subfigure}
\begin{subfigure}[t]{0.19\textwidth}
\includegraphics[width=\textwidth]{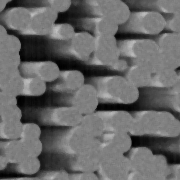}
\end{subfigure}
\begin{subfigure}[t]{0.19\textwidth}
\includegraphics[width=\textwidth]{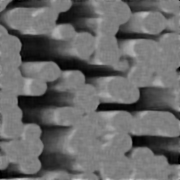}
\end{subfigure}
\begin{subfigure}[t]{0.19\textwidth}
\includegraphics[width=\textwidth]{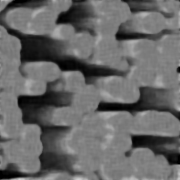}
\end{subfigure}
\begin{subfigure}[t]{0.19\textwidth}
\includegraphics[width=\textwidth]{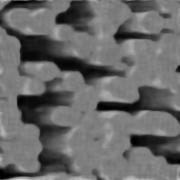}
\end{subfigure}
\begin{subfigure}[t]{0.19\textwidth}
\includegraphics[width=\textwidth]{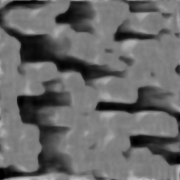}
\end{subfigure}
\begin{subfigure}[t]{0.19\textwidth}
\includegraphics[width=\textwidth]{imgs/FIB/6_yz_90}
\caption*{Ground truth yz}
\end{subfigure}
\begin{subfigure}[t]{0.19\textwidth}
\includegraphics[width=\textwidth]{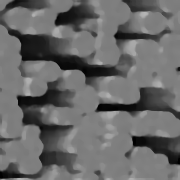}
\caption*{9 yz}
\end{subfigure}
\begin{subfigure}[t]{0.19\textwidth}
\includegraphics[width=\textwidth]{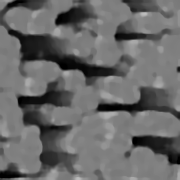}
\caption*{12 yz}
\end{subfigure}
\begin{subfigure}[t]{0.19\textwidth}
\includegraphics[width=\textwidth]{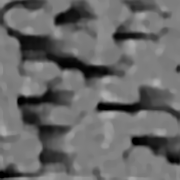}
\caption*{15 yz}
\end{subfigure}
\begin{subfigure}[t]{0.19\textwidth}
\includegraphics[width=\textwidth]{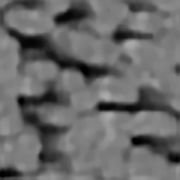}
\caption*{18 yz}
\end{subfigure}
\caption{Slices of the $W_2^2$-regularized (top) and $L^2$-TV reconstructions (bottom) of the high resolution 3D-FIB-SEM-image for the different initial spacings. Top: xy section (SEM imaging plane), bottom: yz section.}
\label{fig_FIB_REM}
\end{figure*}

We set the number of scales to $L=2$.
As a comparison, we use $L^2$-TV using the operator $\hat f$ as forward model.
The resulting error measures are given in Table~\ref{tab_FIB_REM_results}. 
Furthermore, slices of the reconstructions using $W_2^2$-regularization and $L^2$-TV are shown in Figure~\ref{fig_FIB_REM}.
We observe that also in this example the $W_2^2$-regularized reconstruction is the best one visually and in terms of the quality measures.

\begin{table}
\begin{center}
\begin{tabular}{c|c|ccc}
LR spacing&Error measure&Tricubic&$L^2$-TV&$W_2^2$-regularized\\\hline
\multirow{3}{*}{9 nm}&PSNR&$27.12$&$30.90$&$\mathbf{31.29}$\\
&Blur effect&$0.4887$&$0.4758$&$\mathbf{0.4530}$\\
&LPIPS&$0.0869$&$0.0901$&$\mathbf{0.0348}$\\\hline
\multirow{3}{*}{12 nm}&PSNR&$23.15$&$28.61$&$\mathbf{29.09}$\\
&Blur effect&$0.5369$&$0.4995$&$\mathbf{0.4717}$\\
&LPIPS&$0.1945$&$0.1076$&$\mathbf{0.0707}$\\\hline
\multirow{3}{*}{15 nm}&PSNR&$20.18$&$26.23$&$\mathbf{27.33}$\\
&Blur effect&$0.5804$&$0.5588$&$\mathbf{0.5111}$\\
&LPIPS&$0.3039$&$0.2222$&$\mathbf{0.1093}$\\\hline
\multirow{3}{*}{18 nm}&PSNR&$18.56$&$24.11$&$\mathbf{25.74}$\\
&Blur effect&$0.6290$&$0.6159$&$\mathbf{0.5229}$\\
&LPIPS&$0.3975$&$0.3766$&$\mathbf{0.1406}$
\end{tabular}
\end{center}
\caption{PSNR, blur effect and LPIPS value of the high-resolution reconstruction of the simulated 3D-FIB-SEM images using different methods. The original image has a blur effect of $0.4400$.}
\label{tab_FIB_REM_results}
\end{table}

\subsection{Real FIB-SEM images}\label{sec_fib_rem_real}
\noindent
As a real data example we consider a pair of FIB-SEM images of a porous zirconium dioxide filtration membrane produced by spin coating \cite{Carter2007CeramicMS}. The sample is imaged using a Carl Zeiss Crossbeam NVision 40 Field Emission Scanning Electron Microscope (FE-SEM) with integrated advanced tomography package Atlas 5 3-D-Tomography. To reduce charging during SEM imaging of the poorly electrically conductive ZrO2, the sample is sputtered with gold. An additional 1-2 µm platinum layer is applied locally by FIB deposition to smooth the rough porous surface.

The high resolution image contains 649 $\times$ 452 $\times$ 161 voxels with cubic voxels of edge length 10 nm. The size of the low resolution image is 355 $\times$ 272 $\times$ 52 voxels with a voxel edge length of 20 nm.

\paragraph{Forward Operator.}
As in the previous subsection, the forward operator is unknown.
As the synthetic data from Subsection \ref{sec_fib_rem} simulates the 
FIB-SEM imaging process, we will use the forward operator from Subsection \ref{sec_fib_rem}
for superresolution from the 12 nm image to the 6 nm image.

\begin{figure*}[!t]
\centering
\begin{subfigure}[t]{0.3\textwidth}
\includegraphics[width=\textwidth]{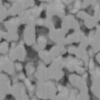}
\caption*{Low resolution xy}
\end{subfigure}
\begin{subfigure}[t]{0.3\textwidth}
\includegraphics[width=\textwidth]{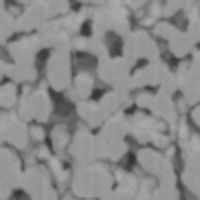}
\caption*{Tricubic xy}
\end{subfigure}
\begin{subfigure}[t]{0.3\textwidth}
\includegraphics[width=\textwidth]{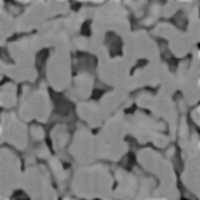}
\caption*{$W_2^2$-regularized xy}
\end{subfigure}
\begin{subfigure}[t]{0.3\textwidth}
\includegraphics[width=\textwidth]{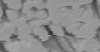}
\caption*{Low resolution yz}
\end{subfigure}
\begin{subfigure}[t]{0.3\textwidth}
\includegraphics[width=\textwidth]{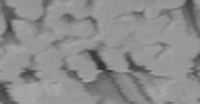}
\caption*{Tricubic yz}
\end{subfigure}
\begin{subfigure}[t]{0.3\textwidth}
\includegraphics[width=\textwidth]{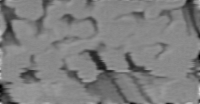}
\caption*{$W_2^2$-regularized yz}
\end{subfigure}
\caption{Slices of the reconstructions of the high-resolution 3D-FIB-SEM-image (spacing 10 nm) for the real data. Top: xy section (SEM imaging plane), bottom: yz section.}
\label{fig_FIB_REM_real}
\end{figure*}

\paragraph{Results.}
We set the number of scales to $L=2$.
Single slices of the resulting reconstruction using the $W_2^2$-regularization and the tricubic interpolation are shown in in Figure~\ref{fig_FIB_REM_real}.
We observe that the $W_2^2$-regularization is sharper than the interpolation. Indeed, the blur effect of the interpolation is given by $0.5107$, while the blur effect of the $W_2^2$-regularized reconstruction is given by $0.4344$.
As no ground truth image is available, the other quality measures cannot be applied in this example.

\section{Conclusions}\label{sec_conclustions}
\noindent
In this paper, we introduced a Wasserstein-$2$ patch prior for image superresolution, which penalizes the Wasserstein-$2$
distance of the patch distribution in the reconstruction to the patch distribution in some reference image.
The minimization of the arising objective functional can be done via a gradient descent scheme based on 
\cite{HLPR2021}.
Finally, we demonstrated the performance of the new prior on 2D and 3D material images.
In particular, we have shown that the method is applicable also in real-world applications.
In case of the FIB-SEM images, the forward operator estimated from the synthetic data also yields reasonable results for the real data. An investigation of the generality and robustness of this approach will be subject of future research.

\section*{Acknowledgements}
\noindent
This study has been carried out with financial support from the French Research Agency through the GOTMI project (ANR-16-CE33-0010-01) and the German Research Foundation (DFG) with\-in the project STE 571/16-1.
The data from Section~\ref{sec_material_data} has been acquired in the frame of the EU Innovative Training Network MUMMERING (Grant Number 765604) at the beamline TOMCAT by A. Saadaldin, D. Bernard, and F. Marone Welford. We acknowledge the Paul Scherrer Institut, Villigen, Switzerland for provision of synchrotron radiation beamtime at the TOMCAT beamline X02DA of the SLS. 
Furthermore, we thank Diego Rold\'an from the TU Kaiserslautern for the generation of the synthetic FIB-SEM images from Section~\ref{sec_fib_rem} and S\"oren H\"ohn (Fraunhofer IKTS) for providing the real FIB-SEM data. All data sets discussed in Sections~\ref{sec_fib_rem} and \ref{sec_fib_rem_real} were obtained in project REPOS [03VP00491/5] funded by the German Federal Ministry of Education and Research. 
 
\bibliographystyle{abbrv}
\bibliography{references}

\end{document}